\documentclass{article}
\usepackage{spconf,amsmath,graphicx}
\usepackage{arydshln}
\usepackage[T1]{fontenc}
\usepackage[utf8]{inputenc}
\usepackage{threeparttable}

\title{QA4QG: Using Question Answering to Constrain Multi-Hop Question Generation}
\name{Dan Su, Peng Xu {\normalfont and} Pascale Fung}
\address{
  Department of Electronic and Computer Engineering\\
  Center for Artificial Intelligence Research (CAiRE)\\
  The Hong Kong University of Science and Technology, Clear Water Bay \\
  {\tt dsu@connect.ust.hk,pxuab@connect.ust.hk,pascale@ece.ust.hk }}

\begin{document}
\maketitle
\begin{abstract}
Multi-hop question generation (MQG) aims to generate complex questions which require reasoning over multiple pieces of information of the input passage. Most existing work on MQG has focused on exploring graph-based networks to equip the traditional Sequence-to-sequence framework with reasoning ability. However, these models do not take full advantage of the constraint between questions and answers. Furthermore, studies on multi-hop question answering (QA) suggest that Transformers can replace the graph structure for multi-hop reasoning. Therefore, in this work, we propose a novel framework, QA4QG, a QA-augmented BART-based framework for MQG. It augments the standard BART model with an additional multi-hop QA module to further constrain the generated question. Our results on the HotpotQA dataset show that QA4QG outperforms all state-of-the-art models, with an increase of 8 BLEU-4 and 8 ROUGE points compared to the best results previously reported. Our work suggests the advantage of introducing pre-trained language models and QA module for the MQG task.
\end{abstract}

\section{Introduction}
\label{sec:intro}
Question generation (QG) is the task of automatically generating a question from a given context and an answer. It can be an essential component in education systems~\cite{yao2018teaching}, or be applied in intelligent virtual assistant systems to make them more proactive. It can also serve as a complementary task to boost QA systems~\cite{tang2017question}.

Most of the previous works on QG focus on generating the SQuAD-style single-hop question, which is relevant to one fact obtainable from a single sentence. Recently, there has been a surge of interest in QG for more complex multi-hop question generation, such as HotpotQA-style questions ~\cite{yang2018hotpotqa,pan-etal-2020-semantic,su-etal-2020-multi,10.1145/3366423.3380114,wang-etal-2020-answer,gupta2020reinforced,sachan2020stronger}. This is a more challenging task that requires identifying multiple relevant pieces of information from multiple paragraphs, and reasoning over them to fulfill the generation. Due to the multi-hop nature of MQA task, different models ~\cite{pan-etal-2020-semantic,su-etal-2020-multi, 10.1145/3366423.3380114,sachan2020stronger} have been proposed to introduce graph-based networks into the traditional Sequence-to-sequence (Seq2Seq) framework to encode the multi-hop information. However, some of the most recent work has shown that the graph structure may not be necessary, and can be replaced with Transformers or proper use of large pre-trained models for multi-hop QA ~\cite{shao2020graph,groeneveld2020simple}. This motivates us to explore Transformer-based architectures for the relational reasoning requirements of the multi-hop QG (MQG) task. 

Another limitation of previous works is that they aim to model $P(\text{Question}|\langle\text{Answer},\text{Context}\rangle)$, and ignore the strong constraint of $P(\text{Answer}|\langle\text{Question},\text{Context}\rangle)$. As suggested by~\cite{tang2017question}, QA and QG are dual tasks that can help each other. We argue that introduction of a multi-hop QA module can also help MQG.

In this paper, we propose \textbf{QA4QG}, a QA-augmented BART-based framework for MQG. We augment the standard BART framework with an additional multi-hop QA module, which takes the reverse input of the QG system (i.e., question $Q$ and context $C~\footnote{The context can be either sentences or paragraphs}$ as input), to model the multi-hop relationships between the question $Q$ and the answer $A$ in the given context $C$. 
QA4QG outperforms all state-of-the-art models on the multi-hop dataset HotpotQA, with an increase of 8 BLEU-4 and 8 ROUGE points compared to the best results reported in previously published work. Our work suggests the necessity to introduce pre-trained language models and QA modules for the MQG task.

\vspace{-10pt}

\section{Related Work}
\vspace{-10pt}


Most previous approaches on MQG have tried to extend the existing Seq2Seq framework for single-hop QG with reasoning ability. One branch of work models text as graph structure~\cite{su-etal-2020-multi, pan-etal-2020-semantic,10.1145/3366423.3380114} and incorporates graph neural networks into the traditional Seq2Seq framework, mostly the encoder. While this graph-based approach is very intuitive, it relies on additional modules such as semantic graph construction, name entity recognition (NER) and entity linking (NEL), which make the whole framework complicated and fragile.

Another branch of work on MQG ~\cite{gupta2020reinforced, wang-etal-2020-answer, xie2020exploring} focuses more on the decoder and aims to augment the Seq2Seq framework with extra constraints to guide the generation.~\cite{gupta2020reinforced} employ multi-task learning with the auxiliary task of answer-related supporting sentences prediction.  ~\cite{wang-etal-2020-answer} integrate reinforcement learning (RL) with answer-related syntactic and semantic metrics as reward. The closest effort to our \textbf{QA4QG} is by ~\cite{xie2020exploring}, who introduce a QA-based reward based on SpanBERT in their RL-enhanced Seq2Seq framework,  to consider the answerability of the generated question.

On the other hand, the most recent work ~\cite{shao2020graph,groeneveld2020simple} has shown the strong capability of simple architecture design with large pre-trained language models for multi-hop QA. Such approaches have outperformed the graph network based methods and achieved comparable performance with state-of-the-art architectures such as \textbf{HGN}~\cite{fang-etal-2020-hierarchical}. This inspires us to explore large pre-trained models for MQG.

\vspace{-0.2in}
\section{Methodology}
\begin{figure}[tp]
    \centering
    \includegraphics[width=0.85\linewidth]{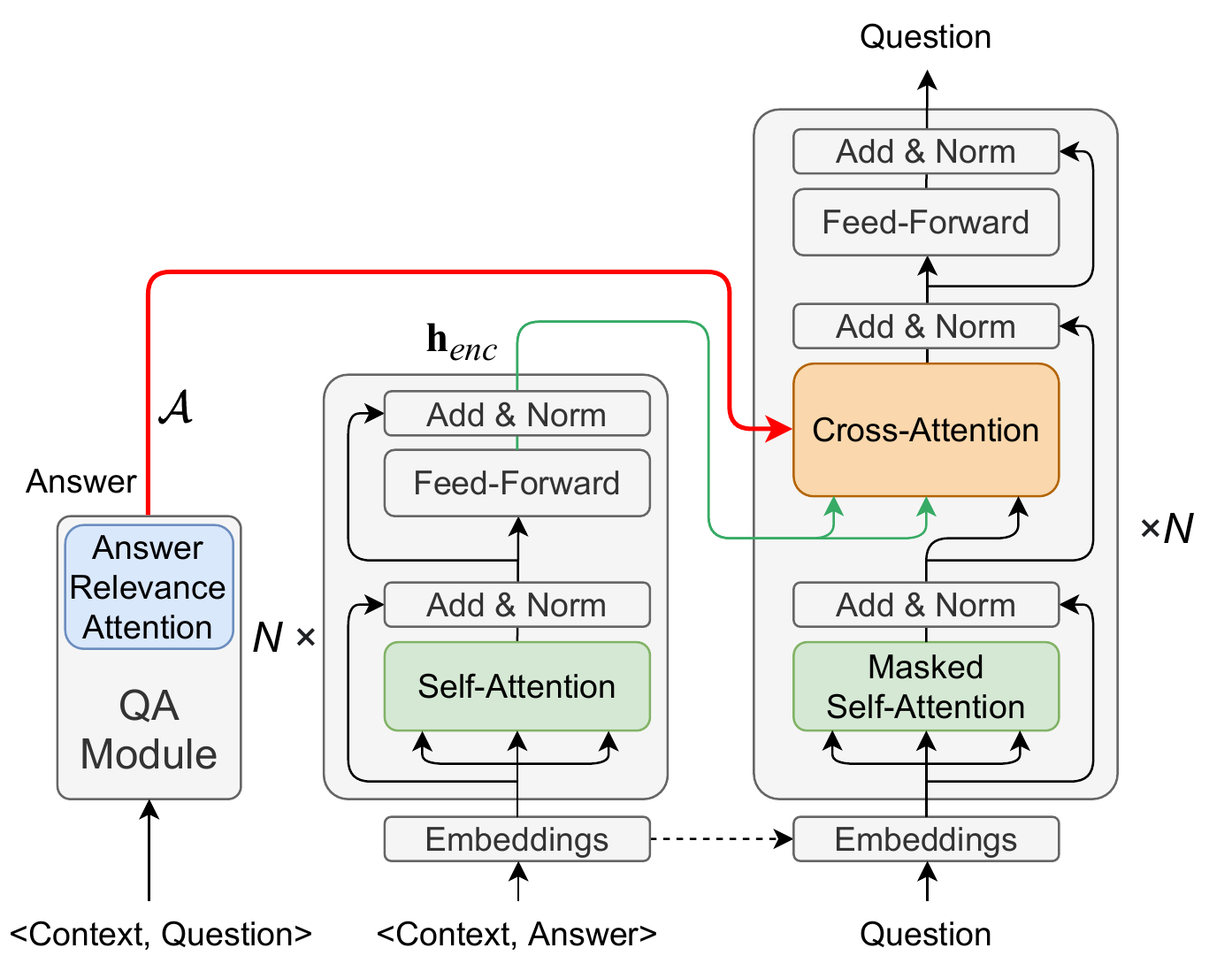}
    \vspace{-10pt}
    \caption{The architecture of our QA4QG. The output of the QA module is used to bias the cross-attention of Transformer decoder.}
    \label{fig:QA4QG}
    \vspace{-0.2in}
\end{figure}

\vspace{-0.15in}
Our framework, QA4QG, consists of two parts, a BART module and a QA module, as shown in Fig ~\ref{fig:QA4QG}. The QA module takes context $C$ and question $Q$ as input, and outputs the probability of each token being the answer. The BART module takes the concatenation of the context $C$ and the answer $A$, together with the output probability from the QA module as input and generates the question $Q$ token-by-token. 
\setlength\doublerulesep{0.35pt}
\begin{table*}[!ht]
\centering
\resizebox{0.80\textwidth}{!}{
\begin{tabular}{l|cccccc}
\hline
\hline
\textbf{Models} & \textbf{BLEU-1} & \textbf{BLEU-2} & \textbf{BLEU-3} & \textbf{BLEU-4} & \textbf{METEOR} & \textbf{ROUGE-L} \\ \hline
\multicolumn{7}{c}{\textit{Encoder Input: Supporting Facts Sentences}} \\ 
\hline
ASs2s-a~\cite{kim2019improving} & 37.67 & 23.79 & 17.21 & 12.59 & 17.45 & 33.21 \\
SemQG~\cite{zhang2019addressing} & 39.92 & 26.73 & 18.73 & 14.71 & 19.29 & 35.63 \\
F + R + A~\cite{xie2020exploring} & 37.97 & - & - & 15.41 & 19.61 & 35.12 \\
SGGDQ (DP)~\cite{pan-etal-2020-semantic} & 40.55 & 27.21 & 20.13 & 15.53 & 20.15 & 36.94 \\
ADDQG~\cite{wang-etal-2020-answer} & 44.34 & 31.32 & 22.68 & 17.54 & 20.56 & 38.09 \\
\hline
QA4QG (\textit{LARGE setting}) & \textbf{49.55} & \textbf{37.91} & \textbf{30.79} & \textbf{25.70} & \textbf{27.44} & \textbf{46.48} \\ 
\hline
\multicolumn{7}{c}{\textit{Encoder Input: Full Document Context}} \\ \hline
MultiQG~\cite{su-etal-2020-multi} & 40.15 & 26.71 & 19.73 & 15.2 & 20.51 & 35.3 \\
GATENLL+CT~\cite{sachan2020stronger}& - & - & - & 20.02(14.5) & 22.40 & 39.49 \\
LowResouceQG~\cite{yu2020low} & - & - & - & 19.07 & 19.16 & 39.41 \\ 
\hline
QA4QG (\textit{BASE setting})  & 43.72&	31.54&	24.47 &	19.68 &	24.55& 40.44 \\
\hdashline
QA4QG (\textit{LARGE setting}) & \textbf{46.45} & \textbf{33.83} & \textbf{26.35} & \textbf{21.21} & \textbf{25.53} & \textbf{42.44} \\
\hline
\hline
\end{tabular}}
\caption{\label{results} Comparison between QA4QG and previous MQG methods on the HotpotQA dataset in different encoder input settings. QA4QG outperforms the best models up to 8 BLEU-4 and 8 ROUGE points.}
\vspace{-0.1in}
\end{table*}

\setlength{\abovedisplayskip}{5pt}
\setlength{\belowdisplayskip}{5pt}
\vspace{-0.1in}
\subsection{BART}
\vspace{-0.05in}
We choose BART as our backbone for Seq2Seq model because of its outstanding performance on many generation tasks~\cite{lewis-etal-2020-bart}. BART is a Transformer-based model that consists of an encoder and a decoder. The encoder encodes the concatenation of the context $C$ and the answer $A$. We denote the encoded final representation of the encoder as $h_{enc}$. Partial structure of the BART decoder is detailed as follow:
\begin{align}
    H_i^a &= \text{MultiHeadAttention}(H_i, H_i, H_i) \\
    H_i^b &= \text{Norm}(H_i + H_i^a) \\
\label{cross-attention}    H_i^c &= \text{MultiHeadAttention}(H_i^b, h_{enc}, h_{enc}),
\end{align}
where $H_i$ is the representation for the $i$-th layer. 
\vspace{-10pt}
\subsection{Answer Relevance Attention}
To model the strong relationships of $P(A | C, Q)$, we propose \textbf{answer relevance attention}, to indicate the answer relevance of each token in context to the target question. Our answer relevance attention can be either soft or hard.
\vspace{-0.08in}
\subsubsection{Soft attention}
\vspace{-5pt}
Soft attention can be employed when the ground truth question is available (e.g., in the training phase), and we propose to use a QA module to derive the answer relevance attention. The QA module takes the concatenation of the context $C$ and question $Q$ as input, and outputs the prediction of the start and end spans of the potential answer in the context. Specifically, it outputs two probability distributions over the tokens in the context: $P^s_{ans}$ and $P^e_{ans}$, where $P^s_{ans}$ / $P^e_{ans}$ is the probability that the $i$-th token is the start/end of the answer span in context $C$. The answer relevance attention score $\mathcal{A}_{soft}$ is calculated via
\begin{equation}
    \mathcal{A}_{soft} = P^s_{ans} + P^e_{ans}
\end{equation}
where $\mathcal{A}_{soft} = \{a_i\}$, $a_i$ denotes the answer relevance of the $i$-th token in context to the question. 

For the QA module of our MQG task, we choose the Hierarchical Graph Network (\textbf{HGN})~\cite{fang-etal-2020-hierarchical} as it achieves the state-of-the-art performance on the HotpotQA dataset. We believe the $\mathcal{A}_{soft}$ generated by the HGN model,\footnote{We use the RoBERTa-large based HGN model, trained on the HotpotQA dataset, and released by the author via https://github.com/yuwfan/HGN} when trained to answer multi-hop question, can naturally learn the answer-aware multi-hop information related to the question inside the context $C$. This information can then complement the BART model for MQG. Note that other QA models can also be adopted in our framework.

\begin{figure}[tp]
\centering
    \includegraphics[width=0.86\linewidth]{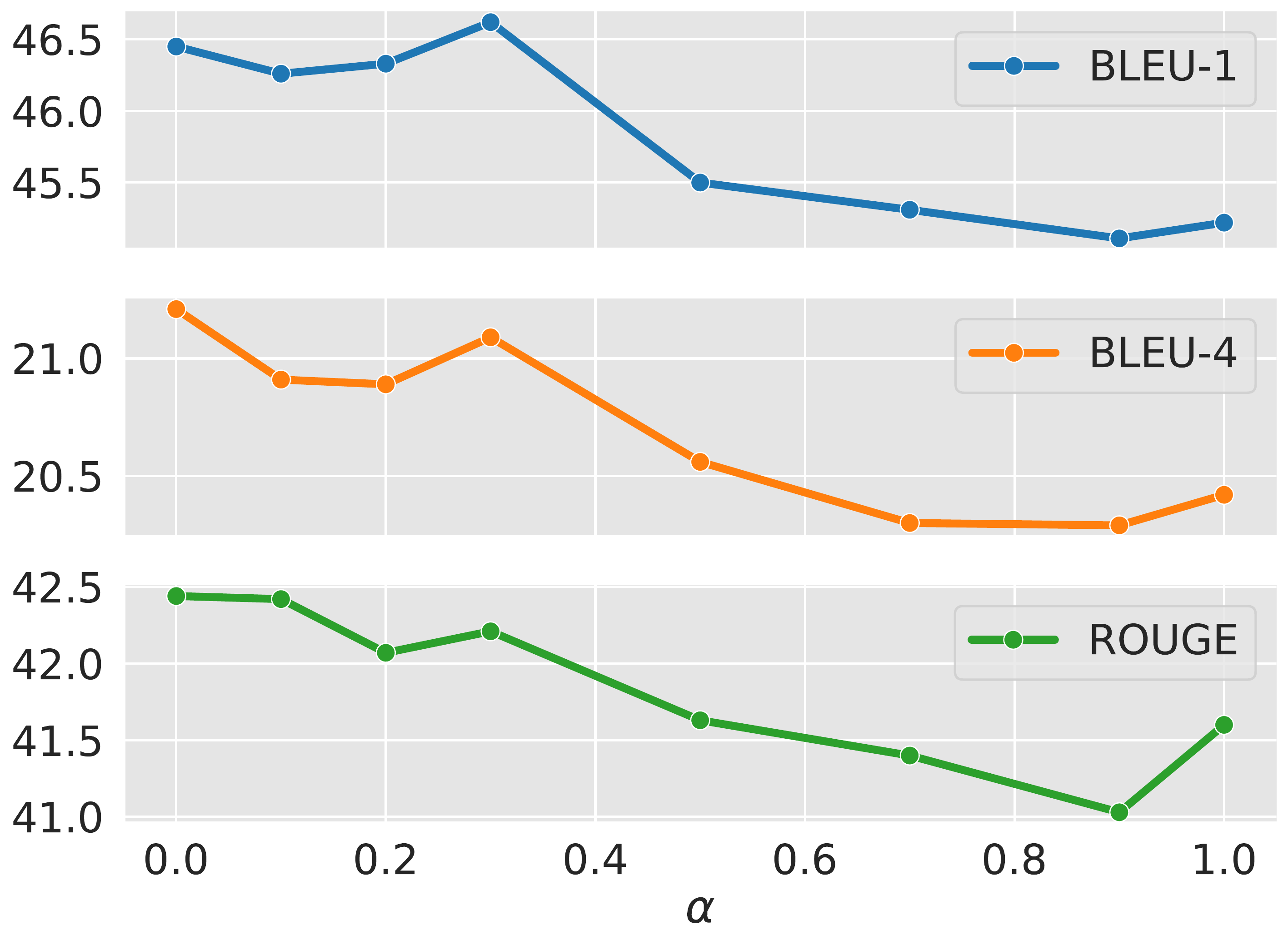}
    \vspace{-0.2in}
    \caption{Ablation study on impact of $\alpha$, with different combinations of the soft attention and hard attention.}
    \label{fig:Ablation}
    \vspace{-0.1in}
\end{figure}

\vspace{-0.08in}
\subsubsection{Hard attention} 
\vspace{-0.08in}
Hard attention can be employed when no question is available (e.g., in the testing phase).  Hard attention is inspired by the answer tagging technique from previous work on single-hop QG. Specifically, we first match the answer span with the context $C$. We then assign a pre-defined score $p_y$ to the matched tokens, and $p_n$ to the remaining tokens, to indicate the binary relevance of each token in the context to the answer (in our work, $p_y=1.0$, $p_n=0.0$). We denote hard attention as $\mathcal{A}_{hard}$.

\vspace{-0.075in}
\subsection{Enhanced Cross-Attention}
\vspace{-0.05in}
We next incorporate the answer relevance attention into the BART model. We propose to bias the original cross-attention sub-layer (i.e., Eq.~\ref{cross-attention}) in each BART decoder layer with $\mathcal{A}$: 
\begin{align}
H_i^{c^\prime} &= \text{softmax}(\frac{H_i^b \, h^T_{enc}}{\sqrt{d_k}} + \mathcal{A})h_{enc} \\
\mathcal{A} &= \alpha \mathcal{A}_{hard} + (1-\alpha) \mathcal{A}_{soft}, \label{eq:combine}
\end{align}



where  $d_k$ is the dimension of keys, and $\alpha$ is the hyper-parameter to mitigate the question disparity between training and testing. Note that each question token shares the same attention score over context tokens. The answer relevance attention can be regarded as a prior knowledge bias for each question over the context. QA4QG is then trained through cross-entropy loss between the generated question and the ground truth.

\section{Experimental Setup}
\vspace{-0.05in}
To evaluate the effectiveness of our QA4QG framework, we conduct experiments on the HotpotQA~\cite{yang2018hotpotqa} dataset, a challenging dataset which contains $ \sim$10k multi-hop questions derived from two Wikipedia paragraphs, and requiring multi-hop reasoning to answer. For fair comparison, we follow the data splits of ~\cite{pan-etal-2020-semantic,wang-etal-2020-answer} to get 90,440 training examples and 6,072 test examples respectively. However, we use the original training data as in HotpotQA, in which each question is paired with two long documents, without pre-precessing. ~\cite{pan-etal-2020-semantic,wang-etal-2020-answer} pre-process the data and only use golden supporting sentences.

\subsection{Training Settings}
We adopt the BART implementations from Huggingface\footnote{https://github.com/huggingface/transformers}, and experiment based on both the \textit{BART-base} and \textit{BART-large} Seq2Seq fine-tuning settings. We run the experiments on single V100 with 16G memory. The maximum source sequence length is set to 512 and 200 respectively, for the full document input and supporting sentences input settings. The training batch size is 6 and 16 respectively for the \textit{QA4QG-base} and \textit{QA4QG-large} model, with gradient accumulation steps of 4. We train all model with maximum 5 epochs. The learning rate is 3e-5. During inference, we use beam search with beam size of 4, and we set the maximum target length to 32 and use the default value of the minimum target length, which is 12, with a length penalty of 1.0.
\vspace{-0.1in}
\subsection{Baselines}
We include the previous work for MQG, and two strong conventional QG models as baselines for comparison:

\textbf{ASs2s-a~\cite{kim2019improving}} proposes to decode questions from an answer-separated passage encoder with a new module termed keyword-net, to utilize the information from both the passage and the target answer.
\textbf{SemQG~\cite{zhang2019addressing}} proposes two semantics-enhanced rewards from question paraphrasing and QA tasks to regularize the QG model for generating semantically valid questions. \textbf{F + R + A~\cite{xie2020exploring}} uses reinforcement learning (RL) and designs three different rewards regarding \textbf{f}leuncy, \textbf{r}elevance and \textbf{a}nswerability, for the MQG task. The answerability reward is generated by an QA model. \textbf{SGGDQ (DP)~\cite{pan-etal-2020-semantic}} uses the supporting sentences as input for MQG. It constructs a semantic-level graph for the input, then uses the document-level and graph level representations to do the generation. \textbf{ADDQG~\cite{wang-etal-2020-answer}} applies RL to integrate both syntactic and semantic metrics as the reward to enhance the training of the ADDQG for MQG task. \textbf{MultiQG~\cite{su-etal-2020-multi}} proposes to integrate graph convolutional neural network with conventional Seq2Seq framework, for the MQG task. They construct an entity graph so that the method can be applied using the full documents as input. \textbf{GATENLL+CT~\cite{sachan2020stronger}} proposes a graph augmented Transformer based framework for MQG. \textbf{LowResouceQG~\cite{yu2020low}} focuses on MQG in a low resource scenario, and proposes to use hidden semi-Markov model to learn the structural patterns from the unlabeled data and transfer this fundamental knowledge into the generation model.

\begin{table}[!t]
\centering
\resizebox{0.85\columnwidth}{!}{%
\begin{tabular}{ll|ccc}
\hline
\hline
\multicolumn{2}{c|}{\textbf{Models}} & \textbf{BLEU-4} & \textbf{METEOR} & \textbf{ROUGE-L} \\ \hline
\multicolumn{2}{l|}{QA4QG-large } & 21.21 & 25.53 & 42.44 \\
 & \textit{w/o}  QA & 19.32 & 24.65 & 40.74\\
 \hline
 \multicolumn{2}{l|}{QA4QG-base} & 19.68 & 24.55 & 40.44 \\ 
 & \textit{w/o} QA & 17.43 & 23.16 & 38.23 \\
 \hline
 \hline
 \multicolumn{2}{l|}{QA4QG-large (sp)} & 25.70 & 27.44 & 46.47 \\
 & \textit{w/o} QA & 25.69 & 27.20 & 46.30 \\ 
\hline
\hline
\end{tabular}%
}
\vspace{-5pt}
\caption{Ablation study on the QA module. The bottom section uses the \textbf{s}upporting \textbf{s}entences (sp) as input.}
\label{tab:ablation}
\vspace{-0.2in}
\end{table}




\vspace{-10pt}
\section{Results and Analysis}
\vspace{-0.05in}
Table~\ref{results} shows the comparison results between our methods and several state-of-the-art MQG models. 

The top section represents using the supporting sentences as input, which is a simplified version of the task. Supporting facts annotations require expensive human labeling and are not always available in the realistic MQG scenario. However, this is the setting used in previous works since their methods can not deal with long documents~\cite{pan-etal-2020-semantic}. We see that in this setting, our QA4QG outperforms previous best results with an absolute gain of around 8 BLEU-4 and 8 ROUGE points.   

We also compare our QA4QG performance on a more challenging setting, using the full document as input. The average length of the context documents is three times the length of the supporting facts in HotpotQA~\cite{sachan2020stronger}. As is evident from the results (bottom section in Table~\ref{results}), QA4QG achieves the new state-of-the-art results in both the \textit{QA4QG-base} and \textit{QA4QG-large} settings.

\begin{figure}[t]
\centering
    \includegraphics[width=1.0\linewidth]{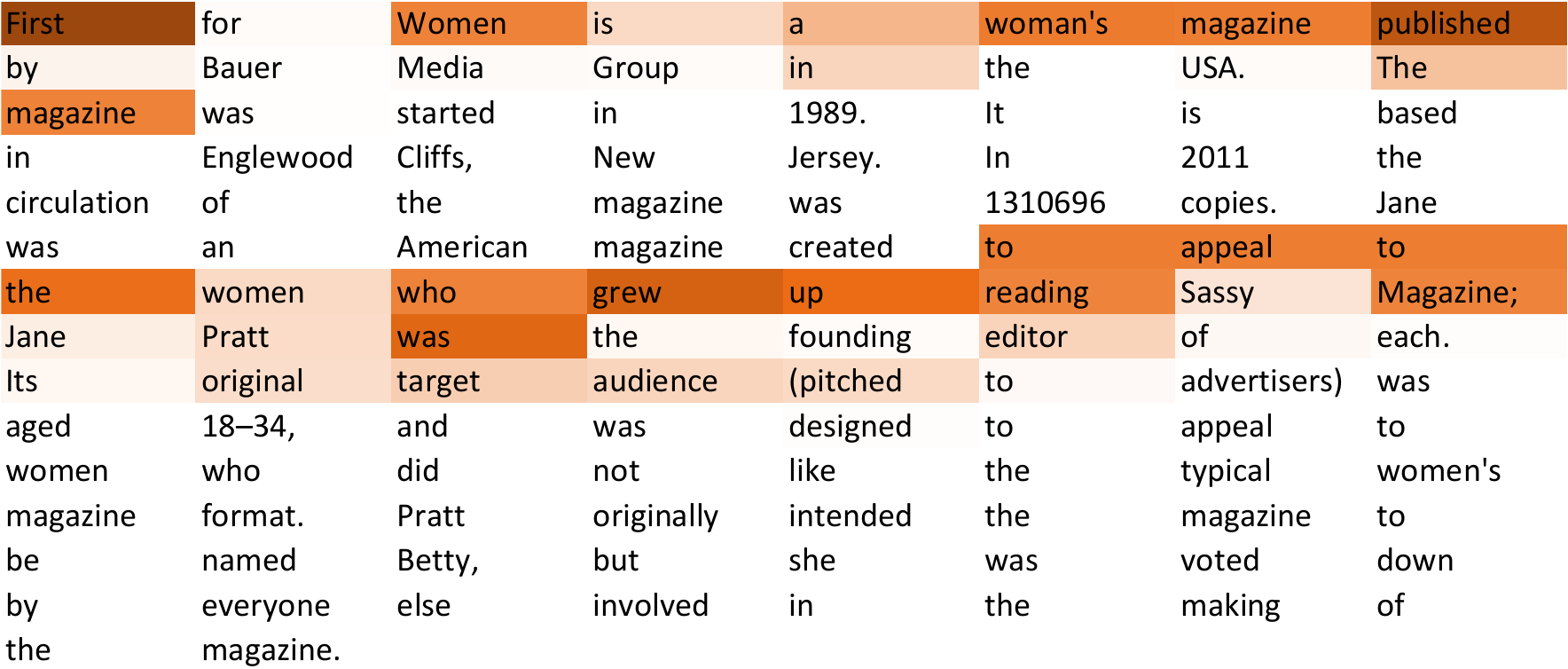}
    \caption{Visualization of soft attention $\mathcal{A}_{soft}$. Darker color represents higher attention weights. For an answer \textit{'yes'}, our $A_{soft}$ emphasizes the multi-hop information related to \textit{'First for Women'} and \textit{'Jane'} in the context, which then constrains the generation model. The target question is \textit{'Are \underline{Jane} and \underline{First for Women} both \underline{women's magazines}?'}.}
    \label{fig:ablation_visualize_attention}
    \vspace{-0.2in}
\end{figure}

\vspace{-0.05in}
\subsection{Ablation Study}
\vspace{-0.05in}
To investigate the answer relevance attention and the QA module, we perform ablation studies. 

As we can see from Table~\ref{tab:ablation}, when we remove the QA module, the performance drops in both the \textit{large} and \textit{base} settings. We also compares when using supporting sentences as input. From the results we see that QA module did not affect the performance as in the full documents setting. This actually matches with our intuition. Since there are only two short sentences as context in the supporting documents setting, it is much easier for the QG model to generate the question, the extra improvement from an QA module may not that large. 

We then study the effect of the hyper-parameter $\alpha$ in Eq.~\ref{eq:combine} with different combinations of soft and hard attention during training. The curves of the three metrics in Fig.~\ref{fig:Ablation} show that, in general, the more $\mathcal{A}_{soft}$, the greater performance improvement QA4QG can achieve. This matches our intuition, since $\mathcal{A}_{soft}$ incorporates the question-related multi-hop information into the context via the QA module, while $\mathcal{A}_{hard}$ only encodes the explicit answer information. The mixture of both when $\alpha = 0.3$ also yields good results, possibly because of the disparity between training and testing, since during testing we only have $\mathcal{A}_{hard}$. 

We visualize the attention weights of $\mathcal{A}_{soft}$ of an example from the dataset in Fig.~\ref{fig:ablation_visualize_attention}. As we see, the $\mathcal{A}_{soft}$ emphasis on the sentence that contains the multi-hop information \textit{'First for Women'} and \textit{'Jane'} in the context, which then constrains the generation model.






\vspace{-0.05in}
\section{Conclusion}
\vspace{-0.05in}
In this paper, we propose a novel framework, QA4QG, a QA-augmented BART-based framework for MQG. It is the first work to explore large pre-trained language models for MQG and takes advantage of an additional Multi-hop QA module to further constrain the question generation. Our results on the HotpotQA dataset show that QA4QG outperforms all state-of-the-art models, with an increase of 8 BLEU-4 and 8 ROUGE points compared to the best results previously reported. Our work suggests the advantage of introducing pre-trained language models and QA modules for the MQG task.


\bibliographystyle{IEEEbib}
\bibliography{strings}

\end{document}